%% file: SFRM_CRC.tex
\documentclass[10pt,twocolumn,letterpaper]{article}

\usepackage{cvpr}
\usepackage{times}
\usepackage{epsfig}
\usepackage{graphicx}
\usepackage{amsmath}
\usepackage{amssymb}
\usepackage{amsmath,amsfonts,amssymb,amsthm,epsfig,epstopdf,array}
\usepackage{subfigure}

\theoremstyle{plain}

\theoremstyle{definition}
\newtheorem{defn}{Definition}[section]

\newtheorem{exmp}{Example}[section]

\theoremstyle{remark}

\usepackage[linesnumbered,noline,ruled]{algorithm2e}

\SetAlCapSkip{1em}
\makeatletter
\newcommand{\removelatexerror}{\let\@latex@error\@gobble}
\makeatother
\SetKwInput{KwInput}{Input}
\SetKwInput{KwOutput}{Output}
\usepackage{float}

\usepackage{hyperref}
\graphicspath{pics/}
\cvprfinalcopy % *** Uncomment this line for the final submission

\def\cvprPaperID{2723} % *** Enter the CVPR Paper ID here

\ifcvprfinal\fi
\begin{document}

%%%%%%%%% TITLE
\title{ Structure from Recurrent Motion: From Rigidity to Recurrency}

\author{
Xiu Li$^{1,2}$ \hspace{0.2in}
Hongdong Li$^{2,3}$ \hspace{0.2in}
Hanbyul Joo$^2$ \hspace{0.2in}
Yebin Liu$^1$ \hspace{0.2in}
Yaser Sheikh$^2$\\\\
Tsinghua University$^1 $\ \hspace{0.1in}
Carnegie Mellon University$^2 $\ \hspace{0.1in}
Australian National University$^3$ }

\maketitle
%\thispagestyle{empty}

%%%%%%%%% ABSTRACT FINAL VERSION
\begin{abstract}
This paper proposes a new method for Non-Rigid Structure-from-Motion (NRSfM) from a long monocular video sequence observing a non-rigid object performing recurrent and possibly repetitive dynamic action.  Departing from the traditional idea of using linear low-order or low-rank shape model for the task of NRSfM, our method exploits the property of shape recurrency (i.e., many deforming shapes tend to repeat themselves in time). We show that recurrency is in fact a {\em generalized rigidity}.  Based on this, we reduce NRSfM problems to rigid ones provided that certain recurrency condition is satisfied. Given such a reduction, standard rigid-SfM techniques are directly applicable (without any change) to the reconstruction of non-rigid dynamic shapes. To implement this idea as a practical approach, this paper develops efficient algorithms for automatic recurrency detection, as well as camera view clustering via a rigidity-check. Experiments on both simulated sequences and real data demonstrate the effectiveness of the method.  Since this paper offers a novel perspective on re-thinking structure-from-motion, we hope it will inspire other new problems in the field.
\end{abstract}
%methods for Non-Rigid SFM (NRSFM), our method does not assume that the non-rigid deformation of the shape of the object  follow certain linear shape or trajectory models (which are  limiting or restrictive in real applications).  

% (SFM)-- a central and classic problem in computer vision research.  We propose a new and original method for reconstructing the 3D shapes of a temporally-deforming  non-rigid object from its multiple images taken from different viewpoints at different time.   

%WE propsoe a new method for NRSFM.   Instead, we leverage on the property fact  observation that temporally-varying shapes of the dynamic object are ofter {\em recurrent} in time, \ie, the shape of the object  (observed at one image frame) tends to repeat itself in time in another frame (,namely,  re-occurring in in time).   Recurrent motions are in fact very common in the physical world, for example, walking pedestrian, running animal,  waving flag,  rotating car wheel, \etc.  We show how to reduce non-rigid SFM problem for this wide class of non-rigid deformable objects to a rigid-SFM.  

   %==================== figure-on next page. 
\begin{figure*}[th!]
\begin{center}
  %%%% \includegraphics[width=0.4\linewidth]{kanadeskewedsymmetry.png} 
   %\hspace{0.85cm}
   \includegraphics[width=\linewidth]{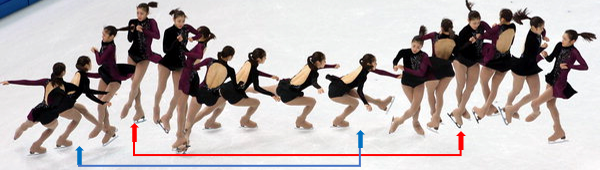}    
\end{center}
 \caption{This composite slow-motion picture of `figure-skating' clearly illustrates the basic idea of our non-rigid SfM method.  Despite the skater's body poses kept changing dynamically over time, there were moments when she struck (nearly) identical posture, e.g., as indicated by the two red arrows and two blue arrows. Using a pair of such recurrent observations- albeit distant in time, one can reconstruct the 3D pose (shape) of the skater at that time instants, by using only standard rigid-SfM techniques.}
\label{fig:1}
\end{figure*}
%============ next page

\input{intro.tex}
%%% move to later sections by hongdong, and remian there \input{relatedwork.tex}
\input{modeling.tex}

\input{epi.tex}
\input{clustering.tex}
\input{results.tex}

\input{relatedwork.tex}

\input{conclusion.tex}
%===========================================
%%%%%%%%% Acknowledgement FINAL VERSION
%\vspace{-0.35in}
\paragraph{Acknowledgement.} {\footnotesize  We would like to thank reviewers and ACs for their valuable comments. This work was completed when XL was a visiting PhD student to CMU under the CSC Scholarship (201706210160). HL is grateful for YS for his very generous hosting.  HL's work is funded in part by Australia ARC Centre of Excellence for Robotic Vision (CE140100016). YL's research is funded by the National Key Foundation for Exploring Scientific Instrument (2013YQ140517) and NSFC grant (No. 61522111).}
%\section*{Acknowledgement} We would like to thank reviewers and ACs for their valuable comments. This work was completed when XL was a visiting PhD student to CMU under the CSC Scholarship (201706210160). HL is grateful for YS for his very generous hosting.  HL's work is funded in part by Australia ARC Centre of Excellence for Robotic Vision (CE140100016). YL's research is funded by the National Key Foundation for Exploring Scientific Instrument (2013YQ140517) and NSFC grant (No. 61522111).
%=========================
{
\bibliographystyle{ieee}
{\small{\bibliography{egbib.bib}}}
}
\end{document}

%% file: intro.tex
\section{Introduction}   
Structure-from-Motion (SfM) has been a success story in computer vision. Given multiple images of a rigidly moving object, one is able to recover the 3D shape of the object as well as camera locations by using geometrical multi-view constraints. Recent research focus in SfM has been extended to the reconstruction of non-rigid dynamic objects or scenes from multiple images, leading to ``Non-Rigid Structure from Motion" (or NRSfM in short).

Despite remarkable progresses made in NRSfM, existing methods suffer from  serious limitations. Notably, they often assume simple linear models, either over the non-rigid shape~\cite{dai2014simple} or over motion trajectories~\cite{akhter2009trajectory}, or both~\cite{gotardo2011non}. These linear models, while they are useful for characterizing certain classes of deforming objects (e.g, face, human pose, or clothing), are unable to capture a variety of dynamic objects in rapid deformation, which are however common in reality.

%After all, suppose one is provided with a video clip containing a non-rigid object with unknown type of deformation, and is asked to reconstruct the 3D dynamic shapes of the object,  there is in fact no practical way for him to validate whether or not the object has a linear shape model before he actually reconstructs it assuming linear shape model -- a dilemma in itself.   This situation partly explains why existing NRSfM methods have not found wide-spread applications so far.    

This paper presents a new method for non-rigid structure from motion. Contrary to the traditional wisdom for NRSfM, we do not make a linear model assumption. Instead, we describe how to exploit shape {\em recurrency} for the task of non-rigid reconstruction. Specifically, we observe that in our physical world many deforming objects (and their shapes) tend to repeat themselves from time to time, or even only occasionally. In the context of SfM if a shape reoccurs in the video we say it is {\em recurrent}.

% Han: the following is modified by Han. I wanted to point out the relation between recurrence and multiview geometry, which was a bit blurry when I read this introduction. 
% Hongdong: I slightly change it a bit, to smooth out some of the expressions. 
%The underlying principle of our new NRSfM method is
This observation of recurrency enables us to use the existing knowledge of multi-view geometry to reconstruct a shape. Given a video sequence, if one is able to recognize a shape that was seen before, then these two instances of images can be used as a virtual stereo pair of the same {\em rigid} object in a space. Therefore one can simply apply standard rigid SfM techniques to reconstruct a non-rigid object, without developing new methods. For instance, the techniques used in rigid SfM~\cite{hartley2003multiple} such as the use of the fundamental matrix, computing camera poses by the Perspective-n-Point (PnP) algorithm, triangulating 3D points, bundle adjustment, and rigid factorization can be used without modifications. We conducted experiments on both synthetic and real data, showing the efficacy of our method.

%Our method is suited to the cases wherever shape seen in one frame repeats itself in another frame in time.  

% 
%
%
%
%%%%%%%%% BODY TEXT
\section{The Key Insight}

Rigidity is a fundamental property that underpins almost all work in rigid Structure-from-Motion (SfM). We say an object is {\em rigid} if its shape remains constant over time. For this reason, multiple images of the same object, taken from different viewpoints, can be viewed as redundant observations of the same target, making the task of rigid SfM mathematically well-posed and solvable. In contrast, the shape of a non-rigid object changes over time, violating the rigidity assumption and rendering NRSfM ill-posed.
  
In this paper, we show that {\em shape recurrency} is in fact a {\em generalized rigidity}. At first glance, finding rigid-pairs can be thought as a restrictive condition; however, satisfying this condition is far easier. Recurrent motions are ubiquitous in our surroundings, including human's walking, animal's running, leaves' waving, clock's pendulum swaying, car wheels' rotating, and so on. Many human motions such as martial arts, dance, and sport games contain various repetitive motions and patterns. Even dramatic or non-periodic motions can be included, as long as a visual observation is long enough, making it highly probable to revisit a previously-seen scene again. If we are given multiple sequences for the similar human motions, although they are not  exactly the same scenes, it can increase the chances of finding recurrent motions.

This is the key insight of this paper. To further illustrate this idea, consider the example of figure skating in Fig.~\ref{fig:1}, showing a composite (strobe-type) photograph made by fusing multiple frames of slow-motion photos, which vividly captures the dynamic performance of the skater.  Examine carefully each of the individual postures of the skater at different time steps; it is not difficult for one to recognize several (nearly) repeated poses. 

%Our new method was inspired the following:  imagine that if our eyes (or our brain) can have long-term visual memory (say, which lasts for seconds, or minutes, even hours), and if a new shape comes to our eyes, which we recognize as being identical/similar to one of the memorized shapes,  then these two images form a virtual stereo image pair of he same 3D shape, despite captured at vastly different time-steps.  As a result, we can simply apply rigid-SFM technique to reconstruct the shape from these two observations.   This is the simple idea of our new method.  

To apply our idea of reconstructing non-rigid shapes from recurrence, we propose a novel method to formulate NRSfM problem as a graph-clustering problem, which can be solved by a Normalized-Cut~\cite{shi2000normalized} framework. In particular, we build a method to compute the probability representing the rigidity of a shape from images at two different time instances. The final recurrence relations are globally solved considering all connections in the constructed graph.
%Although conceptually our idea sounds simple, implementing it as a practical method requires novel and non-trivial (algorithmic) contributions. Specifically, in this paper we develop a novel method to convert NRSfM problem to graph-clustering problem solvable by Normalized-Cut~\cite{shi2000normalized}.  We will show how to quickly determine the probability that two images are projections of the same rigid shape, as well as how to achieve consistent reconstruction.

%% file: modeling.tex
\section{Problem Formulation and Main Algorithm} 

Consider a non-rigid dynamic 3D object observed by a moving pinhole camera, capturing $N$ images at time steps of $t=1,2,..,N.$  Our task is then to recover all the $N$ temporal shapes of the object, $S(1),S(2),..,S(N)$. To be precise,  the shape of the object at time $t$, $\mathbf{S}(t)$, is defined by a set of $M$ feature points (landmarks) on the object: ${S}(t) = [{X}_{t1}, {X}_{t2},..,{X}_{tM}],$  where ${X}_{ti}$ denotes the homogeneous coordinates of the $i$-th feature point of the object at time $t$. Clearly the ${S}(t)$ is a $4\times M$ matrix.  

Given a pinhole camera with a projection matrix $\mathtt{P}$, a single 3D point ${X}$ is projected on the image at position ${x}$ by a homogeneous equation ${x}\simeq\mathtt{P}{X}$.  For the shape of a temporally deforming object at time $t$, we have $\mathbf{x}(t)\simeq\mathtt{P}(t){S}(t),$ where $\mathbf{x}(t) $ denotes the image measurement of the shape $S(t)$ at time $t$, and $\mathtt{P}(t)$ defines the camera matrix of the $t$-th frame. 

By collecting all $N$ frames of observations of the non-rigid object at time $t=1,..,N$, we obtain the basic equation system for $N$-view $M$-point NRSfM problem:  
\begin{equation}
 \begin{bmatrix} \mathbf{x}(1)\\\mathbf{x}(2)\\\vdots\\\mathbf{x}(N) \end{bmatrix}\simeq
 \begin{bmatrix}
  \mathtt{P}(1) &&&\\ 
& \mathtt{P}(2)&& \\ 
  &&\ddots&  \\ 
 &&& \mathtt{P}(N)   \end{bmatrix}\cdot
\begin{bmatrix}S(1)\\S(2)\\\vdots\\S(N)  \end{bmatrix}.
\end{equation}
%===============

\begin{defn} [Rigidity]
Given two 3D shapes ${S}$ and ${S}'$ with correspondences in a space, we can say that they form a {\em rigid pair} if they are related by a rigid transformation $\mathtt{T}.$  Note that a rigid transformation can be compactly represented by a $4\times4$ matrix $\mathtt{T},$ hence we have: ${S}'= \mathtt{T}{S}, ~\exists\mathtt{T}\in\mathbb{SE}(3).$ \end{defn}
%=\begin{bmatrix}R&\mathbf{t}\\\mathbf{0}& 1\end{bmatrix},$ 
We use  $S\approx{S'}$ to denote that $S$ and $S'$ form a {\em rigid pair}. 

\begin{exmp}[Rigid Object]
The shape of a rigid object remains constant all the time: $S(t)\approx{S}(t'), \forall t\ne t'.$
\end{exmp} 

\begin{exmp}[Periodic Deformation]
A non-rigid object undergoing periodic deformation with period $p$ will return to its previous shape after a multiplicity of periods, leading to  
$S(t)\approx{S}(t+kp),\forall{k}\in\mathbb{N}.$
\end{exmp} 

\begin{exmp}[Recurrent Object]
A shape at time $t$ re-occurs after some $\delta$-time lapse: $S(t)\approx{S}(t+\delta).$
\end{exmp} 

\vspace{1in}
\subsection{Rigidity Check via Epipolar Geometry} 

If two 3D shapes (represented by point clouds) and their exact correspondences are given, checking whether they are rigidly related or not is a trivial task. However, this is not possible in the case of NRSfM where the shapes are not known {\em a priori}. All we have are two corresponding images of the shapes, and the rigidity-test has to be conducted based on the input images only.

In this paper, we use {\em epipolar-test} for this purpose.  It is based on the well-known result of epipolar geometry: if two 3D shapes differ by only rigid Euclidean transformations,  then, their two images must satisfy the {\em epipolar relationship}. Put it mathematically, we have $S\approx{S}'\Rightarrow \mathbf{x'}_i^{\top}\mathtt{F}\mathbf{x}_i=0, \forall i,$ where $\mathtt{F}$ is the {\em fundamental matrix} between the two images for $S$ and $S',$ respectively.   Note that the RHS equation must be verified over all pairs of correspondences of $(\mathbf{x}_i,\mathbf{x}'_i),~\forall i$. 

Also note that satisfying epipolar relationship is only a {\em necessary condition} for two shapes $S$ and $S'$ to be rigid. This is because the epipolar relationship is invariant to any $4\times4$ projective transformation in 3-space.  As a result, it is a weaker condition than the rigidity test, suggesting that even if two images pass the epipolar-test they still possibly be non-rigidly related. Fortunately, in practice, this is not a serious issue, because the {\em odds} that a generic dynamic object (with more than 5 landmark points) changes its shape precisely following a 15-DoF 3D projectivity is negligible. In other words, there is virtually no risk of mistaking.

The above idea of epipolar-test looks very simple.  As such, one might be tempted to rush to implementing the following simple and straightforward algorithm: 
\begin{enumerate} 
\item  Estimate a fundamental matrix from the correspondences using the linear 8-point algorithm;
\item Compute the mean residual error computed by averaging all the point-to-epipolar-line distances evaluated on key points in the image;
\item  If this mean residual error is less than a pre-defined tolerance,  return `rigid', else return `non-rigid'. 
\end{enumerate}

%This part is not clear 
{\bf Unfortunately}, despite the simplicity of the above algorithm, it is however not useful in practice, because of the following two reasons. (1) Ill-posed estimation:  It is well known that linear methods for epipolar geometry estimation are very sensitive to outliers;  a single outlier may destroy the fundamental matrix estimation.  However, in our context, the situation is much worse (than merely having a few outliers). This is because,  whenever the two feature point sets are in fact {\em not} rigidly related,  forcing them to fit to a single fundamental matrix by using any linear algorithm can only yield a {\em meaningless} estimation, subsequently leading to a {\em meaningless} residual errors and unreliable decision.  In short,  fitting all feature points to a single epipolar geometry is ill-posed. Instead, in order to do a proper rigidity-test one must consider the underlying 3D rigid-reconstructability of all these image points. (2) Degenerate cases:  Even if two sets of points are indeed connected by a valid and meaningful fundamental matrix, there is no guarantee that a valid 3D reconstruction can be computed from the epipolar geometry.  For example, when the camera is doing a pure rotation,  there will not be enough disparity (parallax) in the correspondences to allow for a proper reconstruction--because the two cameras have only one center of projection- depth can not be observed.  In such cases, the two sets of images can be mapped to each other by a planar homography, and the fundamental matrix estimations are non-unique. 
\paragraph{Our solution:} We propose a new algorithm for rigidity-test, named ``Modified Epipolar Test", which resolves both of the above issues. First, it uses (minimum) sub-set sampling mechanism to ensure that the estimated two-view epipolar geometries (e.g., fundamental matrices) are meaningful. Second, it adopts model-selection to exclude degenerate cases associate with planar homography.  Detailed Epipolar-Test algorithm will be presented in Section-\ref{sec:modifiedEpipolartest}.    
\subsection{Main Algorithm} 
Given the above rigidity-test is in place,  we are now ready to present the main algorithm of the paper, namely {\em Structure-from-Recurrent-Motion (SfRM)}.
{\small{
\begin{figure}[H]
\removelatexerror
\begin{algorithm}[H]
\KwInput{ $N$ perspective views of a non-rigid shape $S(t),  t=1,..N.$  Choose $K$, i.e., the desired number of clusters. }
\KwOutput{The reconstructed 3D shapes of $S(t)$,$\forall t\in\{1,..,N\}$ up to non-rigid transformations.}
  
\For{$(i= 1,\cdots,N, j=1,\cdots,N)$} {Call {\bf Algorithm 2} (i.e., modified-epipolar-test) to get $A$ matrix whose $(i,j)$-th entry $A(i,j)$ gives the probability that the two images $i, j$ are rigidly related.}

[Clustering] Form a view-graph $G(V,E,A)$ connecting all $N$ views, and the $A$ matrix is used as the affinity matrix. Run a suitable graph clustering algorithm to cluster the $N$ views into $K$ clusters. 

[Reconstruction] Apply any rigid SfM-reconstruction method to each of the $K$ clusters. 

\caption{A high-level sketch of our Structure-From-Recurrent-Motion algorithm}
\end{algorithm}
\end{figure}
}}%small 
Note that the core steps of the algorithm are $A$-matrix computation and graph clustering. It should be also noted that our algorithm only makes use of rigid SfM routines to achieve non-rigid shape reconstruction.

%% file: epi.tex
%=======================
\section{Modified Epipolar Test } 
\label{sec:modifiedEpipolartest}
In this section, we describe our modified epipolar-test algorithm. The output of this algorithm is the probability measuring whether these two images can be the projections of a same rigid shape. As discussed in the previous section, we implement this by checking {\em whether or not these two sets of correspondences are related by a certain {\em fundamental matrix}, and at the same time {\em not} related by any planar homography}.  The latter condition (i.e., excluding homography) is to ensure 3D reconstruction is possible. Our algorithm is inspired by an early work of McReynolds and Lowe for the same task of rigidity-checking \cite{mcreynolds1996rigidity}, however ours is much simpler---without involving  complicated parameter tuning and non-linear refinement. Rigidity-checking was also applied for solving multi-view geometry problems without via camera motion \cite{Licvpr10}. 

We will proceed by presenting our algorithm description first,  followed by necessary explanations and comments.

%%============
\begin{figure}[H]
\removelatexerror
\begin{algorithm}[H]
\KwInput{ Two input images, with $M$ feature correspondences $\{(\mathbf{x}_i,\mathbf{x}'_i) | i=1..M\}$}
\KwOutput{ The probability $P$ that the two images are rigidly related.}
\caption{Modified Epipolar Test algorithm}
\label{Alg:2}
\end{algorithm}

\end{figure} 

\begin{enumerate} 

\item ({\bf Initialization}): Set parameters $\sigma_F,\sigma_H,\tau_F,\tau_H$. 

\item ({\bf Estimate fundamental matrices}):  Sample all possible 8-point subsets from the M points;   Totally there are ${M\choose 8}$ such subsets.  Store them in a list, and index its entries by $k$.

\For {$k=1,\cdots,{M\choose 8}$}
{  \small 
\begin{itemize}
\item Pick the $k$-th 8-point subset,  estimate fund-matrix $F_k$ with the linear 8-point algorithm. 

\item Given $F_k$, compute the geometric (point to epipolar-line) distances for all the $M$ points by $F_k$, i.e., $d_F(\mathbf{x}'_i, F_k\mathbf{x}_i)$.

\item Convert the distances to probability measures by applying Gaussian kernel. Compute the product of all probability measures by:  
\begin{eqnarray} P_F(k) = \prod_{i=1..M}\exp\left(-\frac{d_F^2\left(\mathbf{x}_i', F_k \mathbf{x}_i\right)} {\sigma_F^2}\right), \label{eq:PFprod}\end{eqnarray}

\end{itemize}

}

Find the minimum of all the ${M \choose 8}$ probabilities:   i.e., \begin{eqnarray} P_F = \min_{k\in{M \choose 8}} P_F(k).\label{eq:PFmin}\end{eqnarray}

%
%Overall, we have: \begin{eqnarray}
%P_F^* == \min_{m\in{N \choose 8}} \prod_{i=1..N}&\exp\left(-\frac{d_F^2\left(x_i', F_mx_i\right)} {\sigma_F^2}\right) \label{eq:PF} \end{eqnarray}
%

\item  ({\bf Estimate homography})

Run a similar procedure as above, for homography estimation, via sampling all 4-point subsets $\l\in{M\choose 4}$.  The overall homography probability can be computed by:  \begin{eqnarray}
P_H = \min_{l\in{M\choose 4}} \prod_{i=1..M}&\exp\left(-\frac{d_H^2\left(\mathbf{x}_i', H_l \mathbf{x}_i\right)} {\sigma_H^2}\right). \label{eq:PH} 
\end{eqnarray}

\item ({\bf Compute overall probability}) By now we have both $P_F$, and $P_H$.  Compare them with their respective tolerances $ \delta_F$, and $\delta_H$.

{\small 
 \uIf{ ($P_F\ge\tau_F$) and ($P_H<\tau_H$),}
{Set  $P = P_F (1-P_H),$~\Return $P$.}
\Else
{  Set $P = 0$,~\Return $P$. }
}
\hrulefill 
\end{enumerate}

 \subsection{Why does the algorithm work? }

In Step 3 of the algorithm~\ref{Alg:2}, we sample subsets of the data points, each consists of 8 points, the minimally required number to {\em linearly} fit a fundamental matrix. This way we avoid forcing too many points to fit a single epipolar geometry. If the cameras are calibrated, one could also sample 5 points and use the non-linear 5-point essential-matrix algorithm for better sampling efficiency (e.g. \cite{Li-5point,6pt}). 

Once a fundamental matrix $F_k$ is estimated from an 8-tuple, we evaluate the probability of how likely every other feature points (not in the 8-tuple) satisfies this fundamental matrix.  Assuming this probability is independent for each point, the product of Eq.\ref{eq:PFprod} gives the total probability $P_k$ of how well this $F_k$ explains all the $M$ points.  Exhausting all  ${M\choose 8}$ subsets, we pick the {\bf least} one (in Eq.\ref{eq:PFmin}) as a (i.e., conservative) estimate of the rigidity score. In Step 4, we repeat a similar sampling and fitting procedure for homography estimation. The idea is to  perform {\em model-selection} \cite{torr1997assessment} to filter out degenerate cases.  Finally in Step 5, we report the overall probability (of rigidity-check for the two images) as the product of $P_F$ and $(1-P_H)$ when $P_F$ is sufficiently high (i.e., $\ge\tau_F$) and $P_H$ is sufficiently low (i.e., $<\tau_H$); otherwise report `0'. In summary, our algorithm provides a way to estimate the rigidity-score defined by the worst-case goodness-of-fitting achieved for all tentative fundamental matrices for each 8-tuple, while at the same time our algorithm favors the case hardly explained by a homography.

%
%A solution to this problem has been proposed by Torr et al. [155]. The methods will try to fit different models to the data and the one explaining the data best will be selected. The approach is based on an extension of Akaike's information criterion [1] proposed by Kanatani [64]. It is outside the scope of this text to describe this method into details. Therefore only the key idea will briefly be sketched here.
%%
%
%In this case the fundamental matrix (corresponding to a 3D scene and more than a pure rotation), a general homography (corresponding to a planar scene) and a rotation-induced homography are computed. Selecting the model with the smallest residual would always yield the most general model.
\subsection{ How to speed up the computation?}
Our Algorithm~\ref{Alg:2} can be computationally expensive due to its exhaustive subset enumeration step (Step 3). For example, when $M=100$, ${M\choose 8}$ gives a large number of 186 billions. 

Below we will show that one can almost safely replace the enumeration step with a randomized sampling process with much fewer samples, yet at little loss of accuracy. Specifically, we only need to replace the first line (of ``For $k\in[1,{M\choose 8}]$...")  in Step-3 with ``Randomly sample minimal 8-tuples for $k\le{K}$ times..". 

Suppose there are about $e$ proportion of valid subsets (i.e., $e$ is the inlier ratio).  By `valid' we mean this 8-tuple gives a rise to a good epipolar geometry which explains all data points well enough.  
Then the odds (i.e., probability) of picking a valid 8-tuple by only sampling once is $e$, and the odds of getting an outlier is $1-e$.  If one samples $K$ times, then the total odds of getting all $K$ outliers is $(1- e)^K$.  Finally, the odds of getting at least one valid estimation is $p=1-(1-e)^K$.  This predicted odds can be very high in practice, suggesting that even a small number of random samples suffices. Note that this proof is akin to the probability calculation used in RANSAC, one can refer to \cite{torr1997development} for details.

%% file: clustering.tex
\section{View Clustering and Block Reconstruction}

For a given video sequence containing $N$ views, we construct a complete {\em view-graph} $G(V,E,A)$ of $N$ nodes  where each corresponds to one view.  $E$ denotes the set of edges, and $A$ the affinity matrix in which $A(i,j)$ measures the similarity between node-$i$ and node-$j$.  

After Step 3 of Algorithm 1, we have obtained an $N\times N$ matrix $A$.  We will use this $A$ as the affinity matrix of our view-graph. 

Fig.~\ref{fig:Pdemo} shows visualizations of example $A$s that characterizes different types of dynamic movements of objects in videos, showing periodic motion, recurrent motion, and rigid motion respectively. Fig.~\ref{fig:Pdemo}, bright colors in the matrix indicate at which views a particular shape re-occurred.

%From these figures, one can easily discern that, the picture from left to right each corresponds to periodic motion, recurrent motion,  and rigid motion respectively. 
\begin{figure}[t]
\setcounter{figure}{1}
\begin{center}
\includegraphics[width=0.3\linewidth]{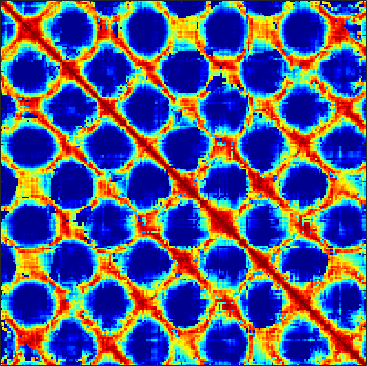} 
\includegraphics[width=0.3\linewidth]{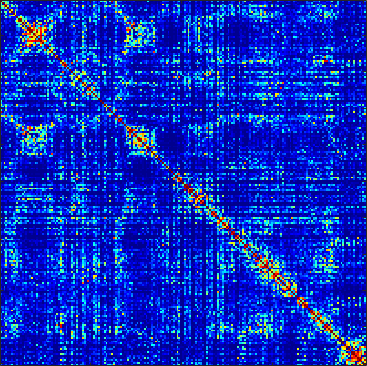} 
\includegraphics[width=0.3\linewidth]{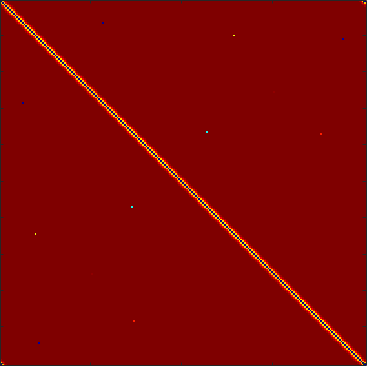} 
\caption{Examples of $A$ matrices: (from left to right), periodic, recurrent, and rigid scenarios.\label{fig:Pdemo}}
\end{center} 
\end{figure}
%====================
%\vspace{-0.2in}
\subsection{Spectral Clustering }
Given a  view-graph $G(V,E,A)$ with the rigidity matrix $A$ as its affinity matrix,  and choose a suitable number $K$ as the intended number of clusters, we suggest to use spectral clustering technique to perform K-way camera view clustering.  If two views are clustered to the same group,  it means the two views are related by a rigid transformation.  

Specifically, we use Shi-Malik's Normalized-Cut~\cite{shi2000normalized} for its simplicity.  The  algorithm goes as follows:  First, compute a diagonal matrix whose diagonal entries are $D(i,i)=\sum_{j}A(i,j) $.  Then, form a Symmetric normalized Laplacian by $L=D^{-1/2}AD^{-1/2}$.  Next, take the least $\log_2K$ eigen-vectors corresponding to the second smallest and higher eigen-values of $L$ and run K-means algorithm on them to achieve $K-$way clustering.  Some examples are given below, in Fig.~\ref{fig:cluster_exp1} and \ref{fig:cluster_exp2}.

\begin{figure}[t]
\begin{center}
\includegraphics[width=0.3\linewidth]{pics/fig2-peroid.png} 
\includegraphics[width=0.3\linewidth]{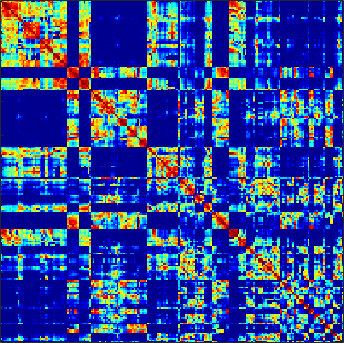} 
\includegraphics[width=0.3\linewidth]{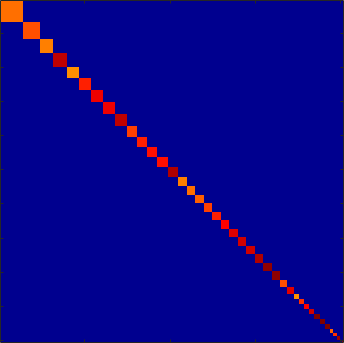} 
\end{center} 
\caption{For periodic motion, with $K=40$ (i.e., one period). From left to right, original $A$ matrix, rearranged $A$ matrix after clustering and clustering membership. }
\label{fig:cluster_exp1}
\end{figure}

\begin{figure}[t]
\begin{center}
\includegraphics[width=0.3\linewidth]{pics/fig2-recc.png} 
\includegraphics[width=0.3\linewidth]{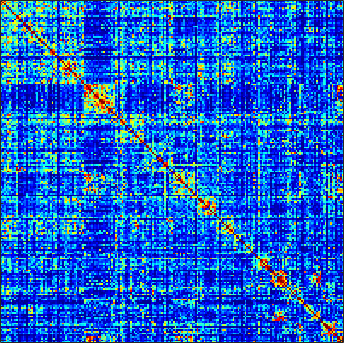} 
\includegraphics[width=0.3\linewidth]{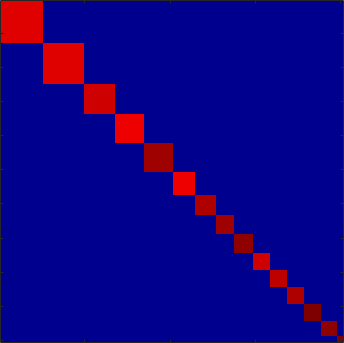} 
\end{center} 
\caption{For general recurrent motion, with $K=25$. From left to right, original $A$ matrix, rearranged $A$ matrix after clustering and  clustering membership.}
\label{fig:cluster_exp2}
\end{figure}

\subsection{Block-wise Rigid Reconstruction}
After the spectral clustering, the $A$ matrix is rearranged to a block-diagonal structure. Each block represents a cluster of views which are rigidly connected, up to an accuracy about the diameter of the cluster.  Therefore, they can be considered as multiple rigid projections of the same shape. Hence any standard rigid-SfM technique can be used to recover the 3D shape. In our experiments specifically, we use incremental bundle adjustment which adds new frames gradually to a local triangulation thread. 

\subsection{Scale Normalization}
As each rigid shape cluster is reconstructed independently, all recovered shapes are up to an ambiguous scale.  To achieve globally consistent reconstruction results, we align the shape scale by normalizing distance between two selected landmarks (e.g., by normalizing the maximum limb-length for human body). 

%% file: results.tex
\section{Results} 
The input of our method is multi-frame feature correspondences, as in other NRSfM methods (e.g., \cite{hartley2008perspective,akhter2009trajectory}). Finding correspondences is a difficult task by itself, especially for non-rigid deforming objects where self-occlusions may happen frequently. In our experiments, for the synthetic data, we assume that the correspondences are provided. For the real data, we use the publicly available OpenPose~\cite{cao2017realtime} library to detect human poses, faces, and hands in sequences.% For other generic objects we use SIFT matching aided with manual correction. %WE also require is dense, for ease of exposition. However, if sparse, our algorithm can be adapted to, with some appropriate modification. as long as block-wise cluster can be found. Camera intrinsic calibration:  We assume the camera is calibrated,  five points can be used for beret stronger lone is a dedicate task, require dedicate effort. 
%=========================================
% hongdong: temporally add the figure back.. 

\subsection{Periodic walking sequence} 
\begin{figure}[t]
\centering
\includegraphics[width=0.45\textwidth]{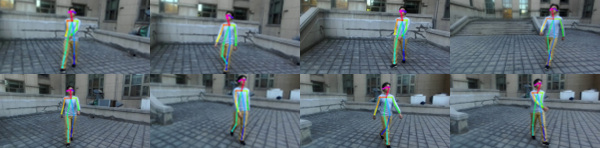}
\caption{A (nearly) periodic walk sequence.}
\label{fig:walk_sim}
\end{figure}

\begin{figure}[t]
\centering
\includegraphics[width=0.15\textwidth]{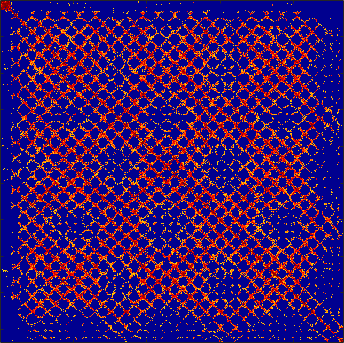}~~
\includegraphics[width=0.15\textwidth]{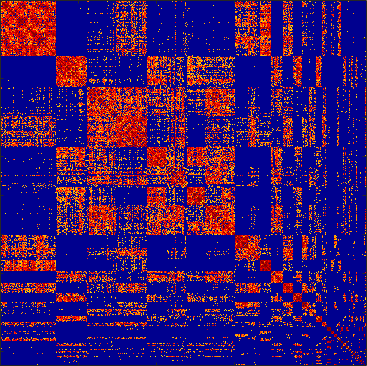}~~
\includegraphics[width=0.15\textwidth]{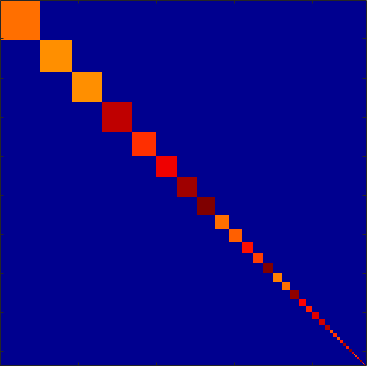}
\caption{Affinity matrices before, and after spectral clustering (\ie N-Cut). The `blocky' structure becomes evident after N-cut. Right: the final view-clustering result.}
\label{fig:walk_cluster}
\end{figure}

\begin{figure}[t]
\centering
\includegraphics[width=0.5\textwidth,trim={{0\linewidth} {0.05\linewidth} {0\linewidth} {0.03\linewidth}},clip]{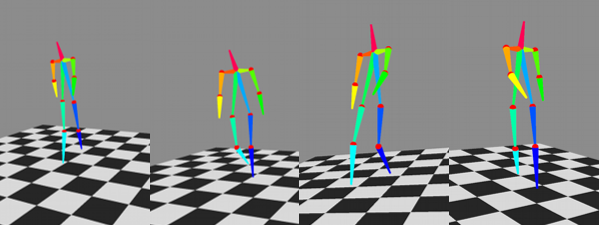}
\caption{3D reconstruction results of the walking sequence.}
\label{fig:walk_shapes}
\end{figure}

This first experiment aims to validate that our Algorithm 1 and 2 work for a real sequence with periodic movements --- which is a special (and simpler) case of recurrent motion. We use a sequence capturing a walking person at a constant speed, where a moving camera is observing this person from different viewpoints, resulting in a nearly periodic sequence. We apply the OpenPose~\cite{cao2017realtime} library to detect 14 landmark points on the person over all 700 frames. Example frames are shown in Fig.~\ref{fig:walk_sim}.  For the entire sequence, the rigidity (i.e., affinity) matrix computed by our Algorithm 2 is shown on the left of Fig.~\ref{fig:walk_cluster}. This figure shows that there exists a strong periodicity, shown as bright bands along the main diagonals. Moreover the period can be readily read out as $p$=40 frames, although our algorithm does not make use of this result. Instead, frames with repetitive shapes are automatically grouped together via view-graph clustering. The middle and the right figure of Fig.~\ref{fig:walk_cluster} show the re-arranged affinity matrix after spectral clustering and the final clustering membership result respectively, where the evident `blocky' structure clearly reveals the grouping.  We then perform a rigid-SfM for all views within each block.  Fig.~\ref{fig:walk_shapes} shows example pose reconstruction results; note the poses are in 3D.  

\begin{figure}[t]
\centering
\includegraphics[width=0.5\textwidth,trim={{0.01\textwidth} {0.01\textwidth} {0.05\textwidth} {0.08\textwidth}},clip]{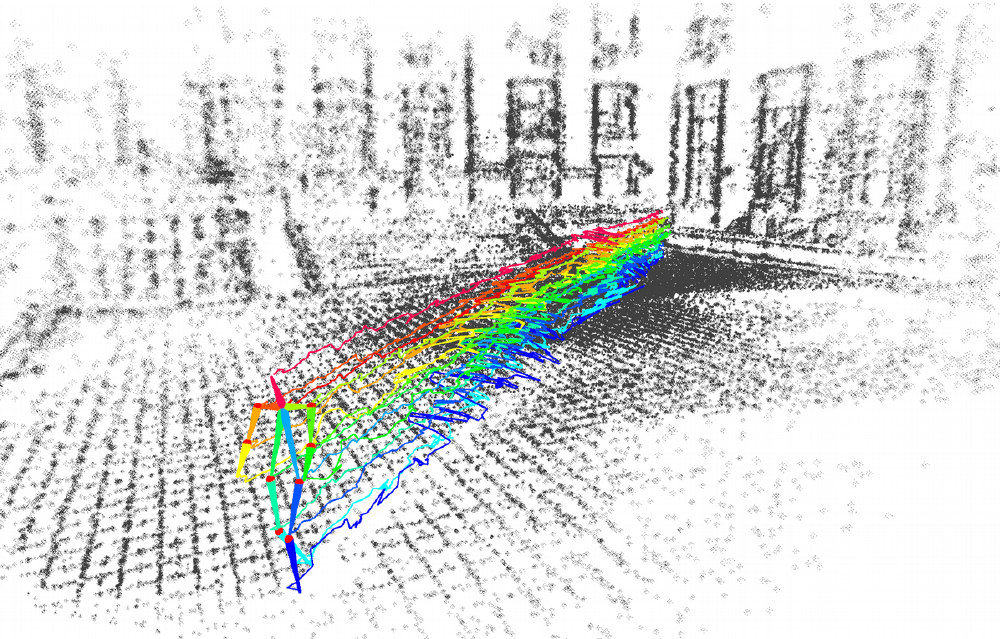}
\caption{Consistent 3D reconstruction of both dynamic foreground object (and temporal trajectories) and a static background scene.} 
\label{fig:walk_withstatic}
\end{figure}

So far, our algorithm has only focused on recovering the non-rigid shape itself, ignoring its absolute pose in the world coordinate frame.  In practice this can be easily fixed, assuming that the ego-motion of each camera view can be recovered by, for example, standard rigid-SfM/SLAM techniques against a stationary background.  We conduct this experiment by first tracking background points, then estimating absolute camera poses relative to the background, followed by Procrustes alignment between the absolute camera poses and each reconstructed human poses. The final reconstruction result, with both background point clouds and human poses and trajectories, is shown in Fig.~\ref{fig:walk_withstatic}. 

\subsection{Recurrent dancing sequence}
This experiment aims to demonstrate the performance of our method on a general ({\em  non-periodic}) video sequence which is likely to contain recurrent movements.  

We choose a solo dancing sequence captured by the CMU Panoptic Studio \cite{Joo_2015_ICCV}. This dataset contains videos from camera arrays. In order to increase the probability of successfully reconstruction, we do not directly use one specific camera, but, instead, extract a time-consecutive video by randomly ``hopping" between different cameras in the dataset, to simulate a video as if captured by a ``monocular camera randomly roaming in space". 

\begin{figure}[t]
\centering
\includegraphics[width=0.156\textwidth]{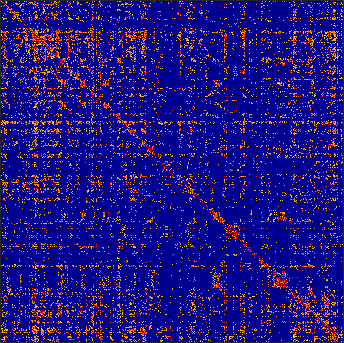}~
\includegraphics[width=0.156\textwidth]{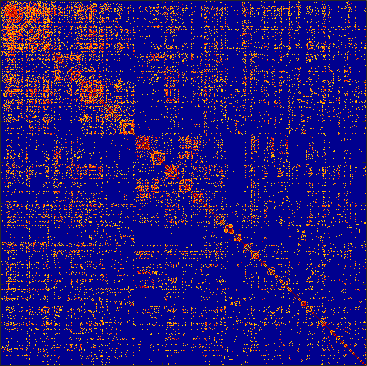}~
\includegraphics[width=0.156\textwidth]{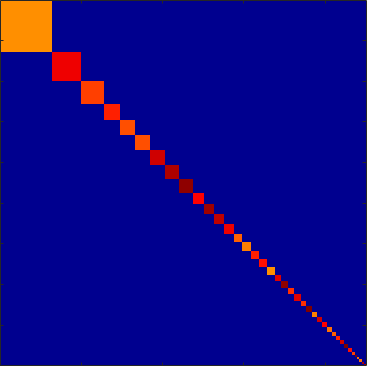}
\caption{The computed original affinity matrix, and the block-wise pattern after cluttering on the CMU dancing sequence. There is no obvious cyclic pattern in the original affinity matrix. After clustering, more clear recurrence patterns are revealed.} 
\label{fig:pano_dance_sim}
\end{figure}

\begin{figure}[t]
\centering
\includegraphics[width=0.45\textwidth]{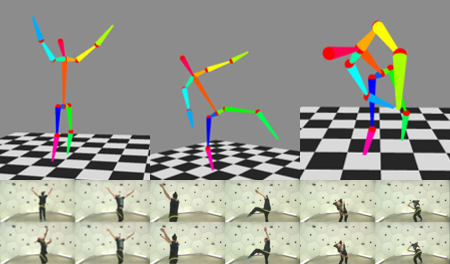}    
\caption{3D reconstruction results on the dance sequence.}
\label{fig:pano_dance_result}
\end{figure}

This dancing sequence is challenging as the motion of the dancer is fast and the dance itself is complicated creating many unnatural body movements.  The computed affinity matrix is shown in Fig.~\ref{fig:pano_dance_sim}, showing that there is no obvious structure. However, after applying our graph-clustering, we can see a clear block-wise pattern (albeit noisy), suggesting that the video indeed contains many recurrent (repetitive) body poses. Example reconstruction results (along with the discovered recurrent frames) are shown in Fig.~\ref{fig:pano_dance_result}.

%c patterns. However, after the clustering stage. The block-wise patterns are revealed. Although  for some blocks they are very noise and contains lots of different pose configurations. That is because in the epipolar-geometry constraint is only neccessary but far from sufficent for rigidity-judgment. We discard all frames with large reprojection errors. And the 

%%----------------------------------------- 
\subsection{Quantitative evaluation}
\begin{figure}[t]
\centering
\includegraphics[width=0.1\textwidth]{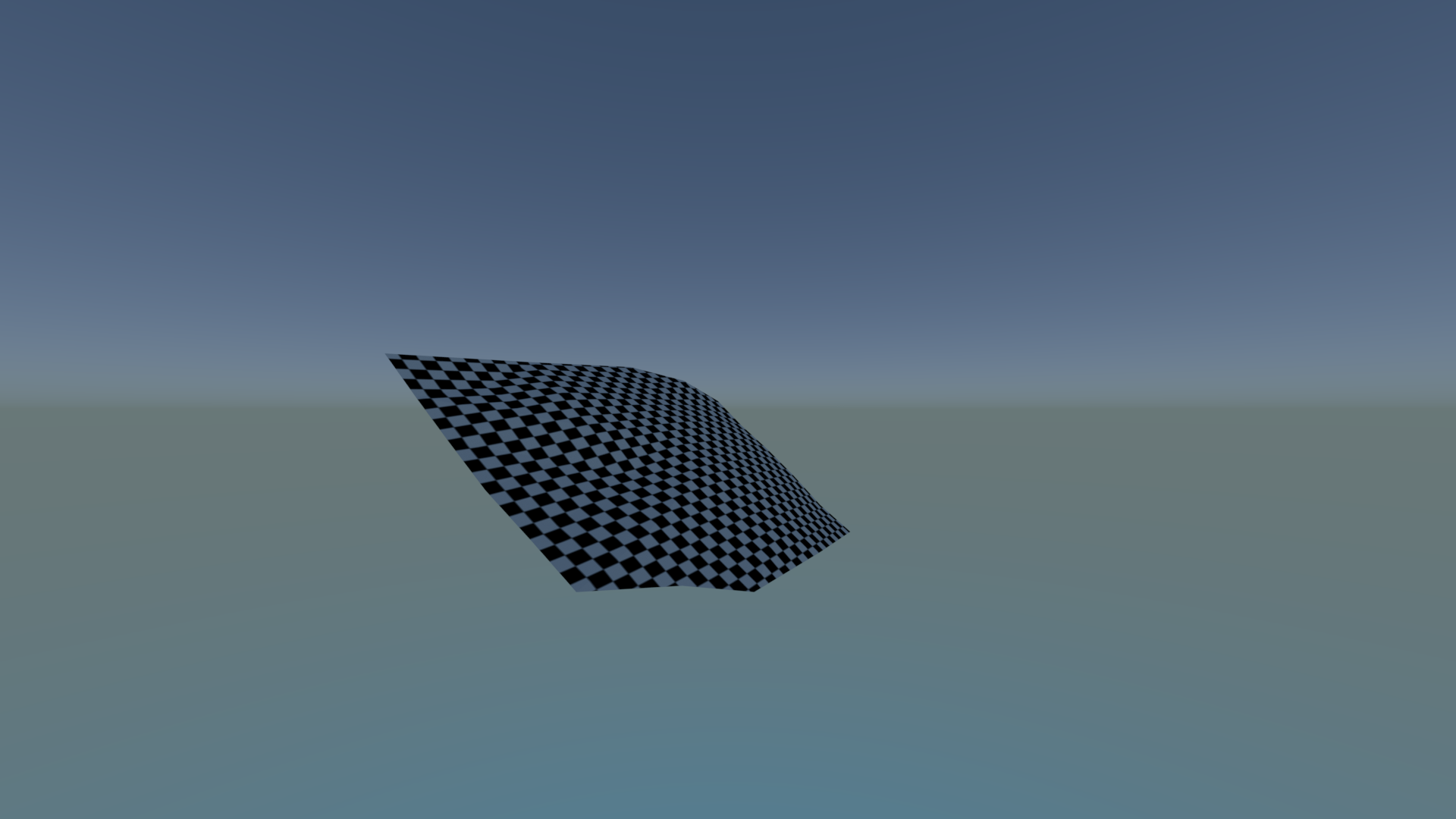}~
\includegraphics[width=0.1\textwidth]{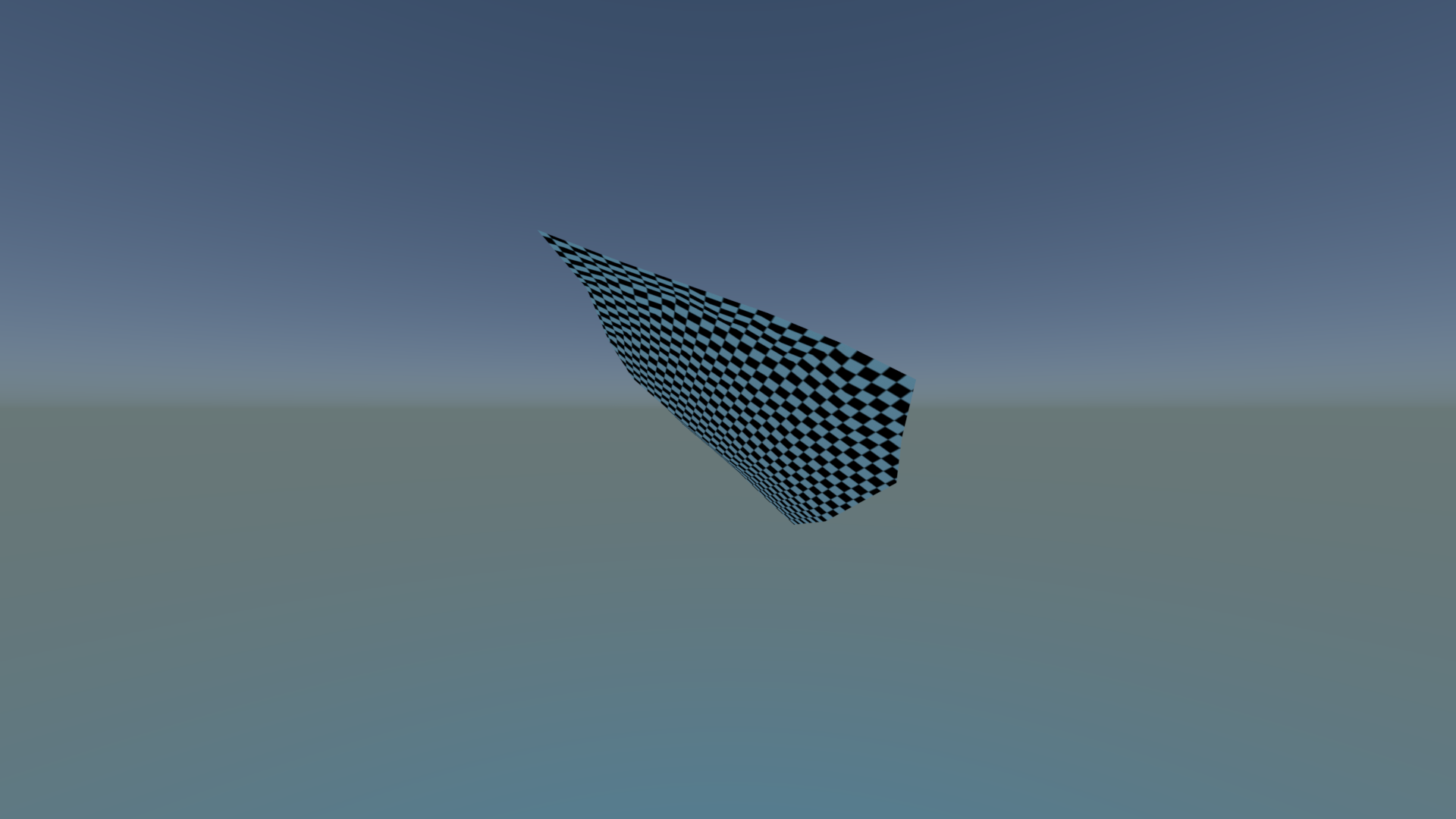}~
\includegraphics[width=0.1\textwidth]{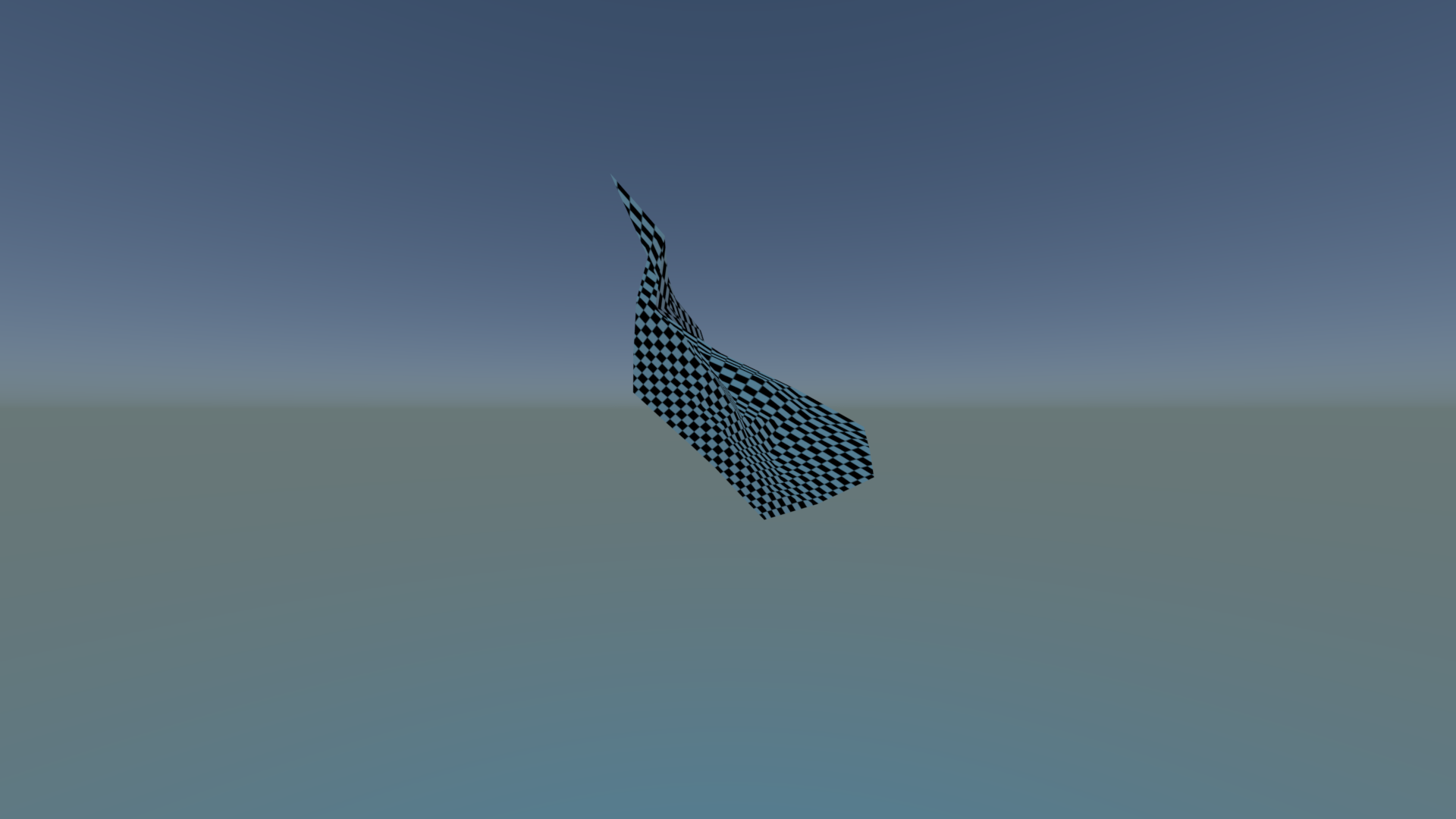}~
\includegraphics[width=0.1\textwidth]{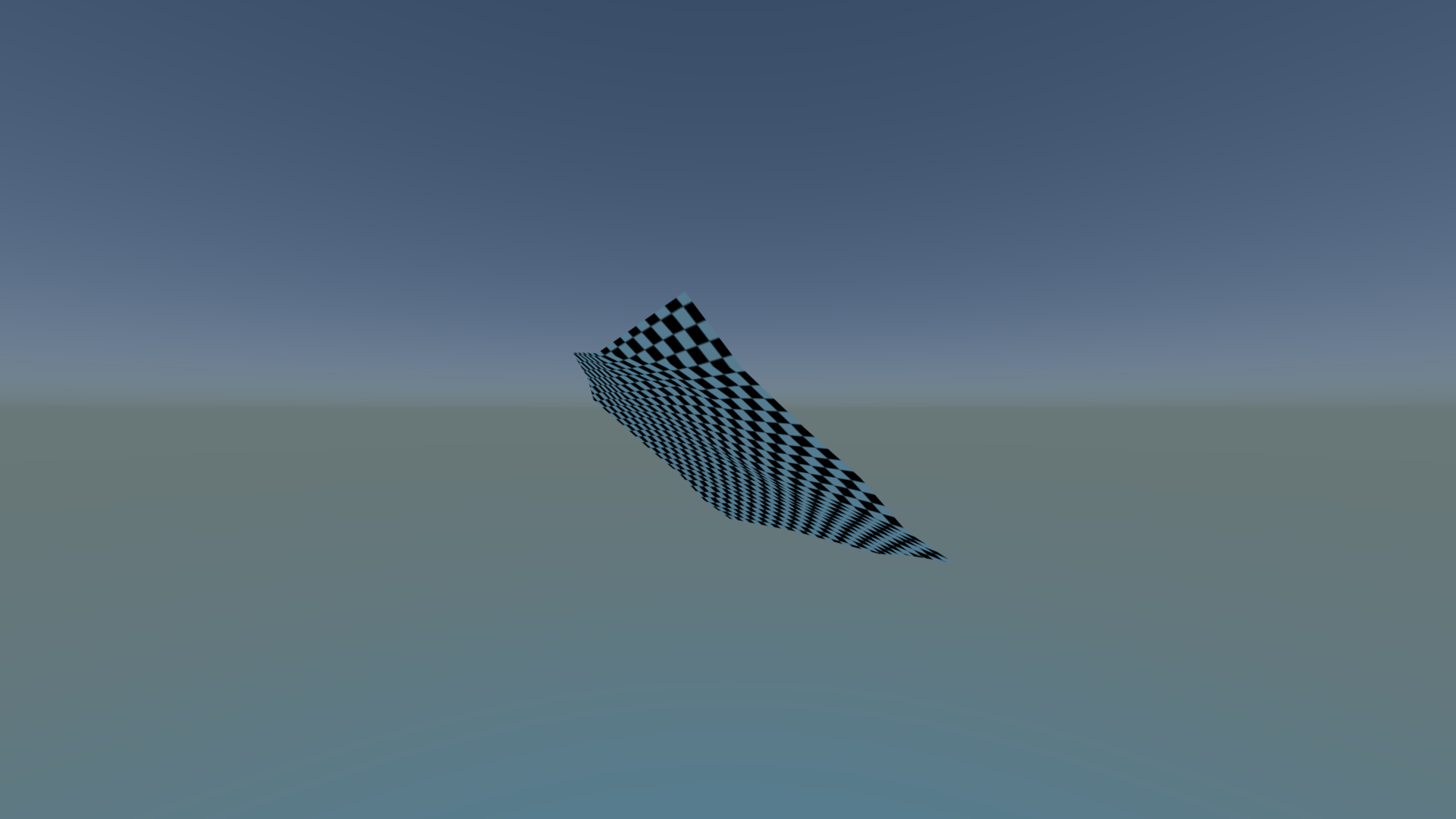}
\includegraphics[width=0.41\textwidth]{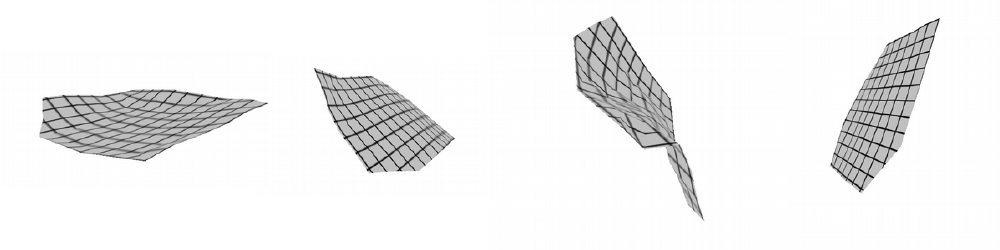}`
\caption{Cloth waving in the wind and our SfRM reconstructions.}
\label{fig:cloth_simu}
\end{figure}

To quantitatively measure the performance of our method, we use the Blender to generate synthetic deformations with recurrence.  We use the flying cloth dataset \cite{white2007capturing} and fold the sequence by several times to mimic recurrency. Camera views are randomly generated. Fig.~\ref{fig:cloth_simu} shows some sample frames of the data. 

In this sequence, all ground-truth (object shape, camera poses) are given. Noises of different levels are added to image planes. Our method  successfully detects recurrency and reconstructs the shape as shown in second row of Fig.~\ref{fig:cloth_simu}.

\begin{figure}[t]
\centering
\includegraphics[width=0.4\textwidth]{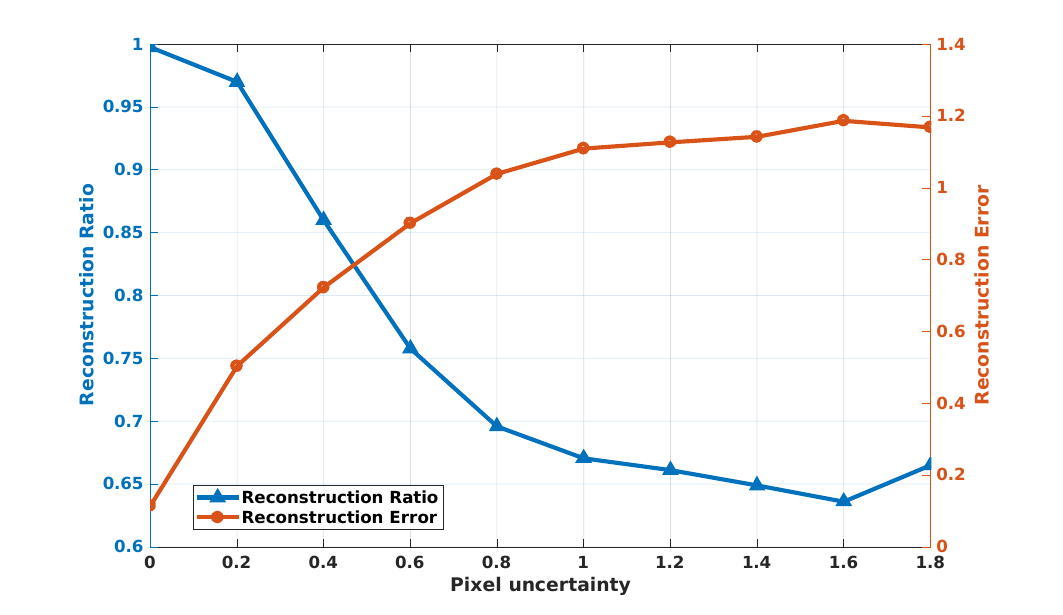}
\caption{SfRM performance at different noise levels. When noise increases, the reconstruction error increases whereas the success ratio falls. This result shows our method handles increasing amount of noises gracefully.}
\label{fig:cloth_noise}
\end{figure}

The reconstruction quality is measured by shape errors after alignment, as well as the portion of successfully reconstructed frames. We evaluate on two criteria at different noise levels. Results are given in Fig.~\ref{fig:cloth_noise}.

\begin{figure}[t]
\centering
\includegraphics[width=0.32\textwidth]{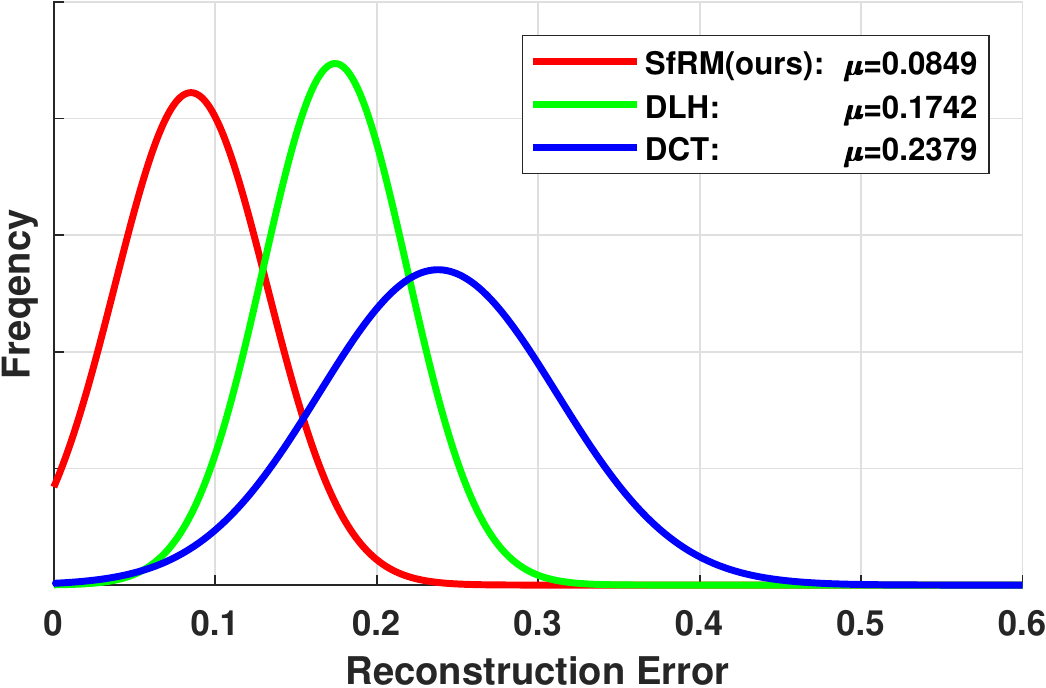}
\caption{Histograms of reprojection errors by different methods. Here we compare our method with shape-basis based method\cite{dai2014simple} and trajectory-basis based method\cite{akhter2009trajectory}. }
\label{fig:cloth_hist}
\end{figure}

We compare our method with other the state-of-the-art template-free NRSfM methods\cite{akhter2009trajectory,dai2014simple}. The result is shown in Fig.~\ref{fig:cloth_hist}. In terms of overall reconstruction accuracy their performances are comparable, while ours is superior for frames exhibiting strong recurrency.

\subsection{Timing}
\begin{figure}[h]
\centering
\includegraphics[width=0.16\textwidth]{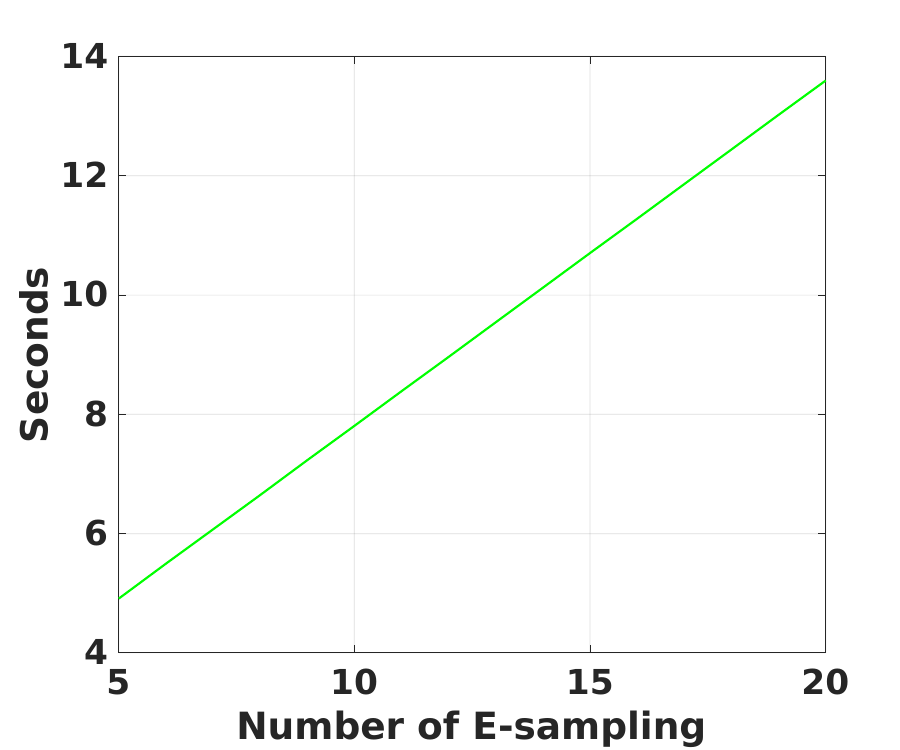}~
\includegraphics[width=0.16\textwidth]{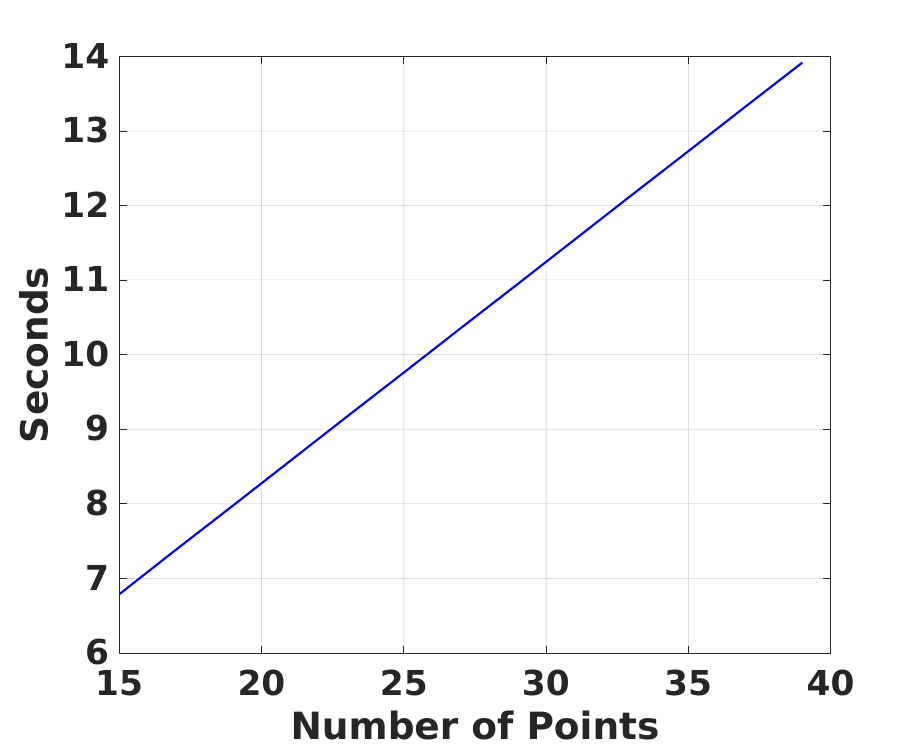}~
\includegraphics[width=0.16\textwidth]{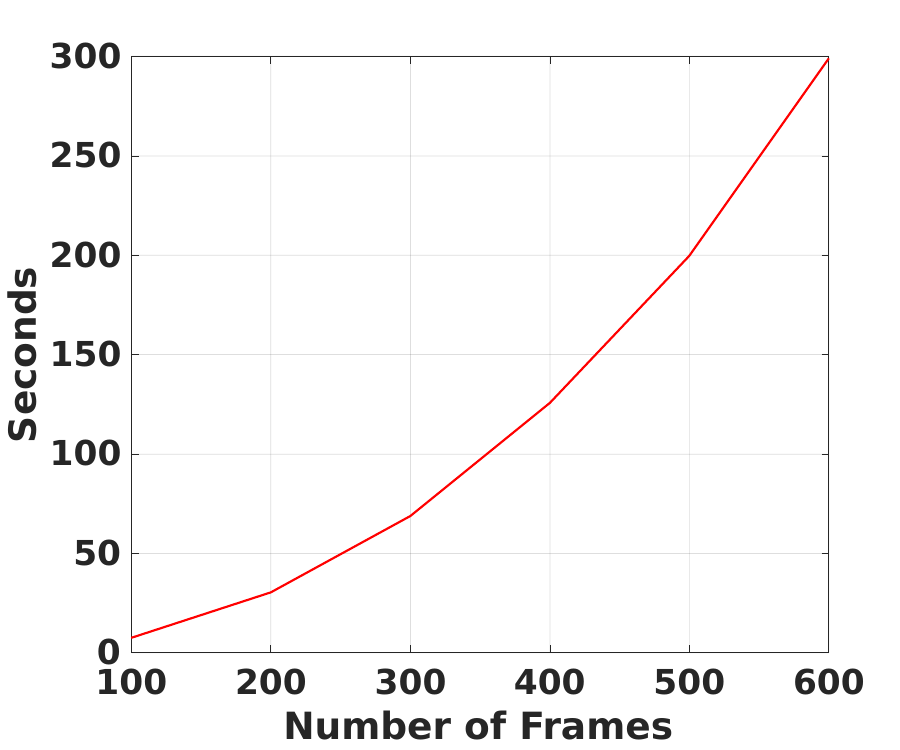}
\caption{ Timing  (in seconds) as a function of the number of random samples, the number of points, and the number of frames. }
\label{fig:runtime}
\end{figure}

Fig.~\ref{fig:runtime} gives the timing results of our SfRM system (excluding rigid reconstruction), showing a clear linear relationship w.r.t. the number of feature points, as well as w.r.t. the number of random samples (in algorithm-2), but is quadratically related to the number of image frames. In our experiments we chose $K$--the number of clusters-- empirically. For future work we would like to investigate how to automatically determine $K$.  

\begin{figure}[h]
\centering
\includegraphics[width=0.35\textwidth]{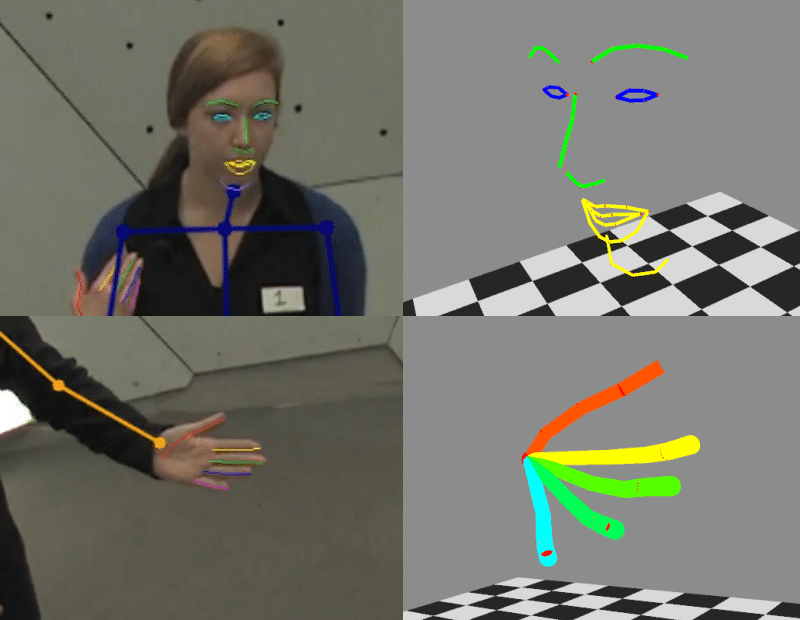}
\caption{Example images of 3D reconstruction of face and hand data.}
\label{fig:face_hand}
\end{figure}

We also test our method on face and hand data captured by the Panoptic Studio. Sample qualitative results are shown in Fig.~\ref{fig:face_hand}. 

%% file: relatedwork.tex
\section{Related work} 
%\textcolor{red} {Xiu: Please ADD the following Five references to bib.}
%Phil Torr, Model SElection work for fundamental matrix, homography... 
%Detecting Irregularities in Cyclic Motion ,
%Steven M  Seitz , Charles R  Dyer
%Periodic Motion Detection and Segmentation via Approximate Sequence Alignment
%Ivan Laptev†, Serge J. Belongie‡, Patrick Pe ́rez† and Josh Wills‡
%Affine Invariant Detection of Periodic Motion
%S. M. Seitz and C. R. Dyer, Proc. Computer Vision and Pattern Recognition Conf., 1994, 970-975.
%Rigidity Checking of 3D Point Correspondences Under Perspective Projection
%Authors:	Daniel P. McReynolds	
%David G. Lowe	
%Published in:
%· Journal
%IEEE Transactions on Pattern Analysis and Machine Intelligence archive
%Volume 18 Issue 12, December 1996
The idea of our SfRM method is rather different from conventional NRSfM approaches. For space reason we will not review the NRSfM literature here but refer interested readers to recent publications on this topic and references therein~\cite{dai2014simple,ParkSMS2010movingpoint,panji, Kumar3DV,bartoli2017dense,CVPR18NRSFM}.  Below, we focus on previous work with similar ideas. 

A cornerstone of our method is the mechanism to detect shape recurrence in a video.  Similar ideas were proposed for periodic dynamic motion analysis \cite{belongie2006structure,ribnick20103d,seitz1994detecting,seitz1994affine}. Our work is specifically inspired by \cite{belongie2006structure,belongieICCV}. However, there are major differences. First, their methods assume strictly periodical motions, and need to estimate the period automatically \cite{cutler2000robust} or manually \cite{belongie2006structure}.  This way, their methods can only handle limited periodical motions such as well-controlled walking and running. In contrast, our method extends to more general cases of recurrent motions, which include both {\em a-periodic}, and re-occurring cases. Moreover, their methods assume a camera to be static, and under the the periodical assumption, the target is not allowed to turn around and has to move (walking or running) on a straight line, capturing only partial surfaces \cite{belongie2006structure} or trajectories \cite{ribnick20103d}. Comparably, our method allows free-form target movements and camera motions. Finally, our method is fully automatic, while their methods rely on significant level of manual interactions. 

Our method can be applied for 3D human pose recovery, therefore it is related to many work in this domain,  \cite{IonescuPOS14Human36,RamakrishnaKS2012pose,PavlakosZDD2016Coarse-to-Fine,VNect2017,gait_AVSS, gait}.  In particular, our method is related to the research directions which try to lift 3D pose from 2D images, e.g. \cite{BogoKLG0B2016SMPL,2d+matching2017,MartinezHRL17simple}.  Earlier work in this direction either requires the integration of knowledge of the bone length of the target \cite{LeeC85}, or human pose and shape space priors \cite{BogoKLG0B2016SMPL}. Although in experiments we use 3D human poses, mainly as exemplar recurrent movements, our method does not take advantage of any category-specific priors. Rather, we treat poses as general point clouds in 3D.  It can be applied to other objects beyond human body.  Another category of work on human pose capture relies on the existence of large-scale pose database for retrieving the most similar pose based on a 3D-2D pose similarity metric \cite{GuptaMLW14,2d+matching2017,Jiang10Millions}. Their performance is heavily depend on the size and quality of the database of specific type of targets, while ours works in general scenarios.  A recent deep learning approach by Martinez et al.~\cite{MartinezHRL17simple} shows that a well-designed network for directly regressing 3D keypoint positions from 2D joint detection showed good performance. However, they rely on large amount of training data of specific class, while ours works without training.

%% file: conclusion.tex
\section{Conclusion} 
We have presented a new method for solving Non-rigid Structure-from-Motion (NRSfM) for a long video sequence containing recurrent motion.  It directly extends the concept of rigidity to recurrency as well as periodicity.  With this new method at hand, one is able to directly use traditional rigid SfM techniques for non-rigid problems.  Key contributions of this work include a randomized algorithm for robust two-view rigidity check, and a view-graph clustering mechanism which automatically discovers recurrent shape enabling the subsequent rigid reconstructions. Finite but adequate experiments have demonstrated the usefulness of the proposed method.    
The method is practically relevant, thanks to the ubiquity of recurrent motions in reality.  One may criticize our method will not work if a shape was only seen for one times.  We admit this is a fair criticism, but we argue that if that happened it would be of little real practical value to reconstruct any shape with such a fleeting nature. Our proposed view-graph and shape-clustering algorithms are examples of unsupervised machine-learning techniques. In this regard, we hope this paper may offer insights that bridge SfM research with learning methods.
%======================